\pdfoutput=1

\documentclass[letterpaper, 10 pt, conference]{ieeeconf}

\IEEEoverridecommandlockouts

\title{\LARGE \bf
ViT-VS: On the Applicability of Pretrained Vision Transformer Features for Generalizable Visual Servoing
}
\author{Alessandro Scherl$^{1,2}$, Stefan Thalhammer$^{2}$, Bernhard Neuberger$^{2}$, Wilfried Wöber$^{2}$, José García-Rodríguez$^{1}$% <-this % stops a space
\thanks{*This work has been submitted to the IEEE for possible publication. Copyright may be transferred without notice, after which this version may no longer be accessible.}% <-this % stops a space
\thanks{$^{1}$The authors are with the with the Department of Computer Technology, University of Alicante, Spain.
 {\tt\small(email: as358@alu.ua.es, jgarcia@dtic.ua.es})}
\thanks{$^{2}$The authors are with the Industrial Engineering Department, UAS Technikum Vienna, Austria.
        {\tt\small (email: \{alessandro.scherl, stefan.thalhammer, bernhard.neuberger,	wilfried.woeber\}@technikum-wien.at})}%
}

\usepackage{booktabs}
\usepackage{xcolor}
\usepackage{amssymb}
\usepackage{amsmath}
\usepackage{graphicx}
\usepackage{wrapfig}
\usepackage{stfloats}
\usepackage{multirow}
\usepackage{gensymb}
\usepackage{rotating}
\usepackage{hyperref}
\usepackage{dirtytalk}
\usepackage{siunitx}

\begin{document}
\maketitle
\thispagestyle{empty}
\pagestyle{empty}

\begin{abstract}

Visual servoing enables robots to precisely position their end-effector relative to a target object. 
While classical methods rely on hand-crafted features and thus are universally applicable without task-specific training, they often struggle with occlusions and environmental variations, whereas learning-based approaches improve robustness but typically require extensive training.
We present a visual servoing approach that leverages pretrained vision transformers for semantic feature extraction, combining the advantages of both paradigms while also being able to generalize beyond the provided sample.
Our approach achieves full convergence in unperturbed scenarios and surpasses classical image-based visual servoing by up to 31.2\% relative improvement in perturbed scenarios.
Even the convergence rates of learning-based methods are matched despite requiring no task- or object-specific training.
Real-world evaluations confirm robust performance in end-effector positioning, industrial box manipulation, and grasping of unseen objects using only a reference from the same category.
Our code and simulation environment are available at: \url{https://alessandroscherl.github.io/ViT-VS/}

\end{abstract}

\section{INTRODUCTION}

Visual servoing (VS) as a visual control strategy allows positioning the robot relative to a target with a single reference~\cite{RAM2006Chaumette,chaumette2007visual}.

This enables executing downstream tasks, such as object tracking and grasping~\cite{chen2022image,de2021dual,puang2020kovis,la2012robotic}.
Generally, VS can be categorized into two approaches: Position-Based Visual Servoing (PBVS) operating on pose differences, and Image-Based Visual Servoing (IBVS), directly utilizing image features.
These features range from geometric primitives and image moments~\cite{chaumette2007visual} to feature descriptors~\cite{hoffmann2006visual} or direct utilization of photometric image data~\cite{ICRA2008Collewet}.

While effective in controlled settings, these classical methods show limited robustness to image perturbations and typically require exact target object instances.
Recent learning-based approaches attempt to address these limitations through different strategies such as pose regression~\cite{ICRA2018Bateux}, velocity regression~\cite{ICRA2021Felton}, learned feature extraction~\cite{CASE2022Adrian}, metric learning~\cite{ICRA2023Felton}, and unsupervised feature learning ~\cite{RAL2022Felton,IROS2022Huh,puang2020kovis,SII2024Quaccia}. 
However, these methods introduce new challenges - they require task-specific training, extensive data generation, or predefined object models, making them difficult to deploy in real-world scenarios where unseen objects and environmental changes are common.

\begin{figure}[!t]
    \centering
    \includegraphics[width=0.99\columnwidth]{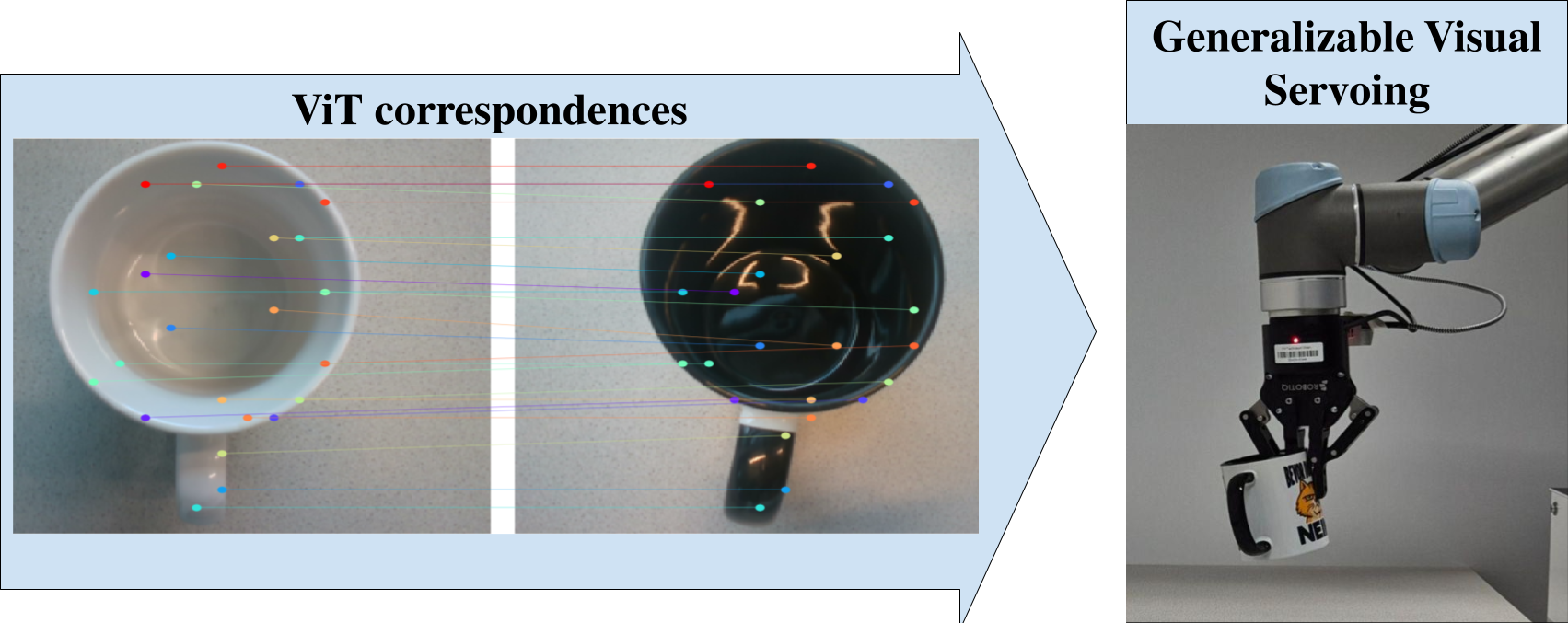}
    \caption{\textbf{ViT-VS category-level object grasping} Left: ViT Correspondence Matching process with white mug as desired image. Right: Successful grasp of the target object.} 
    \label{fig:VitVS_teaser}
    \vspace{-3ex}
\end{figure}
We hypothesize that pretrained Vision Transformers (ViTs)~\cite{ICLR2021Dosovitskiy} effectively combine the advantages of both classical and learning-based VS approaches through semantically robust features.
To validate this, we introduce ViT-VS, a generalizable visual servoing framework combining IBVS with DINOv2~\cite{TMLR2023Oquab} features.
Our approach requires no task-specific training or fine-tuning, and is capable of performing visual servoing robust to image perturbations with high convergence rates. 
Fig. \ref{fig:VitVS_teaser} showcases the ViT feature matching between different object instances (left) and the corresponding successful grasps (right).
A key challenge in utilizing ViTs for VS stems from their rotation invariance property, developed during training for image classification~\cite{zhang2024telling}.
This invariance can cause convergence to incorrect orientations, specifically with in-plane rotations of $\pm 90\degree$ and $180\degree$. 
We address this by aligning the camera based on accumulated feature similarities across different simulated rotations of the current image before initiating the VS control loop.
Additionally, the high computational complexity of ViT image processing leads to suboptimal path lengths during convergence.
To alleviate this behavior we propose a velocity stabilization using an exponential moving average filter.
We evaluate ViT-VS in both simulation and real-world environments, demonstrating:
\begin{itemize}
\item A novel VS approach that is applicable without training or fine-tuning, like classical IBVS approaches, yet achieves the convergence rates and robustness to image perturbations of learning-based approaches. Relative convergence rate improvement is $31.2\%$ as compared to the best classical IBVS, achieving a $100\%$ convergence rate for unperturbed scenarios.
\item Strategies for addressing ViT limitations including compensation for ViTs' rotation invariance and trajectory regularization for reducing convergence length ratios.
\item Industrial box manipulation with $100\%$ success rate $(n=20)$ on a mobile robot with a starting point positioning error of $\pm10cm$.
\item A quantitative evaluation of category-level object grasping with VS used for end-effector positioning. Grasping unseen instances of the categories shoe, toy car, and mug with a mean success rate of $90\%$ $(n=30)$.
\end{itemize}

The remainder of this paper is organized as follows: Section \ref{sec:related_work} reviews related work, Section~\ref{sec:method} presents our novel ViT-VS methodology, Section \ref{sec:experiments_results} details our experimental evaluation and results, and Section \ref{sec:conclusion} concludes the paper with a discussion of our findings.

\section{Related Work}\label{sec:related_work}

This section introduces classical visual servoing schemes, Section \ref{sec:classical_vs}, Section \ref{sec:learning_based_vs} reviews emerging deep learning approaches, and Section \ref{sec:foundation_model_features} discusses the potential of ViT features to advance visual servoing.

\subsection{Classical Visual Servoing}\label{sec:classical_vs}
Classical visual servoing can be broadly classified into IBVS \cite{hutchinson1996tutorial} and PBVS \cite{wilson1996relative}.
These approaches rely on handcrafted geometric features such as points, lines, and moments for robot guidance. 

PBVS requires both camera intrinsic parameters and the object's 3D model, and while it can theoretically achieve global asymptotic stability, it relies heavily on accurate pose estimation.
In contrast, IBVS operates directly in the image space, requiring only camera intrinsic parameters.
IBVS demonstrates robust performance against calibration errors and image noise, though it can only guarantee local asymptotic stability~\cite{RAM2006Chaumette,chaumette2007visual}.
Traditional feature detectors have been utilized for IBVS, with SIFT features performing end-effector positioning~\cite{hoffmann2006visual} and SURF features executing object-specific grasping~\cite{la2012robotic}.
However, while these traditional feature extracting methods offer general applicability, they have been observed to struggle with occlusions, varying illumination, and complex environments~\cite{karami2017image}.
In order to avoid explicit feature extraction Direct Visual Servoing (DVS) was introduced, while achieving lower positioning error compared to classical approaches, it suffers from a limited convergence domain \cite{ICRA2008Collewet}.
These limitations in handling real-world complexities for classical VS strategies have motivated the exploration of learning-based approaches.

\subsection{Learning Based Visual Servoing} \label{sec:learning_based_vs}

In recent years multiple works contributed to visual servoing by utilizing deep learning, overcoming the problematic of occlusion, lighting variations, scene changes, and image perturbations. 
The authors in \cite{ICRA2018Bateux} combine classical PBVS with convolutional neural networks to regress camera pose using synthetically generated datasets for training, demonstrating robust convergence against environmental variations.
Building on pose-based approaches, \cite{ICRA2023Felton} explores deep metric learning by creating a common latent space for camera poses and image representations, incorporating perturbed samples in the training data for enhanced robustness.
Several works utilize siamese networks for visual servoing, in \cite{ICRA2021Felton} directly regressing camera velocity without pose estimation, while the method of~\cite{IROS2022Huh} jointly learns feature extraction and transformation through 3D equivariance constraints for wide-baseline visual servoing.
Alternative feature-based approaches include the work of \cite{RAL2022Felton}, which uses an unsupervised convolutional autoencoder to learn compact image representations that generalize to similar unseen targets.
In the research of \cite{CASE2022Adrian} classical IBVS is combined with neural networks for feature extraction and matching, reaching promising final positioning error though requiring a rendering engine and object model.
Similarly, \cite{puang2020kovis} employs a simulation-to-real transfer approach using object models, learning end-to-end robotic motion and enabling direct deployment on real robots after training purely in simulation.
A recent work~\cite{SII2024Quaccia} replaces pixel brightness with neural network feature representations for DVS, though still facing challenges with illumination variations.
While these learning-based approaches demonstrate significant improvements over classical methods, they either require task-specific training and data generation \cite{ICRA2018Bateux,ICRA2021Felton,RAL2022Felton,IROS2022Huh,ICRA2023Felton} or target object models \cite{CASE2022Adrian,puang2020kovis}, limiting their immediate practical deployment.
This highlights an ongoing challenge in developing more flexible and readily deployable visual servoing solutions.

\subsection{Foundation Model Features} \label{sec:foundation_model_features}

Vision Transformers have recently emerged as powerful architectures for visual tasks, offering robust semantic understanding through self-attention mechanisms~\cite{ICLR2021Dosovitskiy}.
Pre-trained ViTs demonstrate strong zero-shot capabilities and generalization across related object categories~\cite{ICCV2021Caron,CVPR2022Amir}.
While primarily used for classification and detection tasks, their ability to extract general features makes them promising for visual servoing applications.

In this work, we leverage the suitability of ViT features for zero-shot vision tasks. 
By using ViT features for IBVS we combine the advantages of classical and learning-based VS.
The method presented in the next section combines the advantages of classical approaches, i.e. general applicability but no need for offline training or fine tuning, with the advantages of deep learning approaches, i.e. robustness to occlusion, lighting variations and image perturbations.

\begin{figure*}[!t]
    \vspace{1ex}
    \centering
    \includegraphics[width=\textwidth]{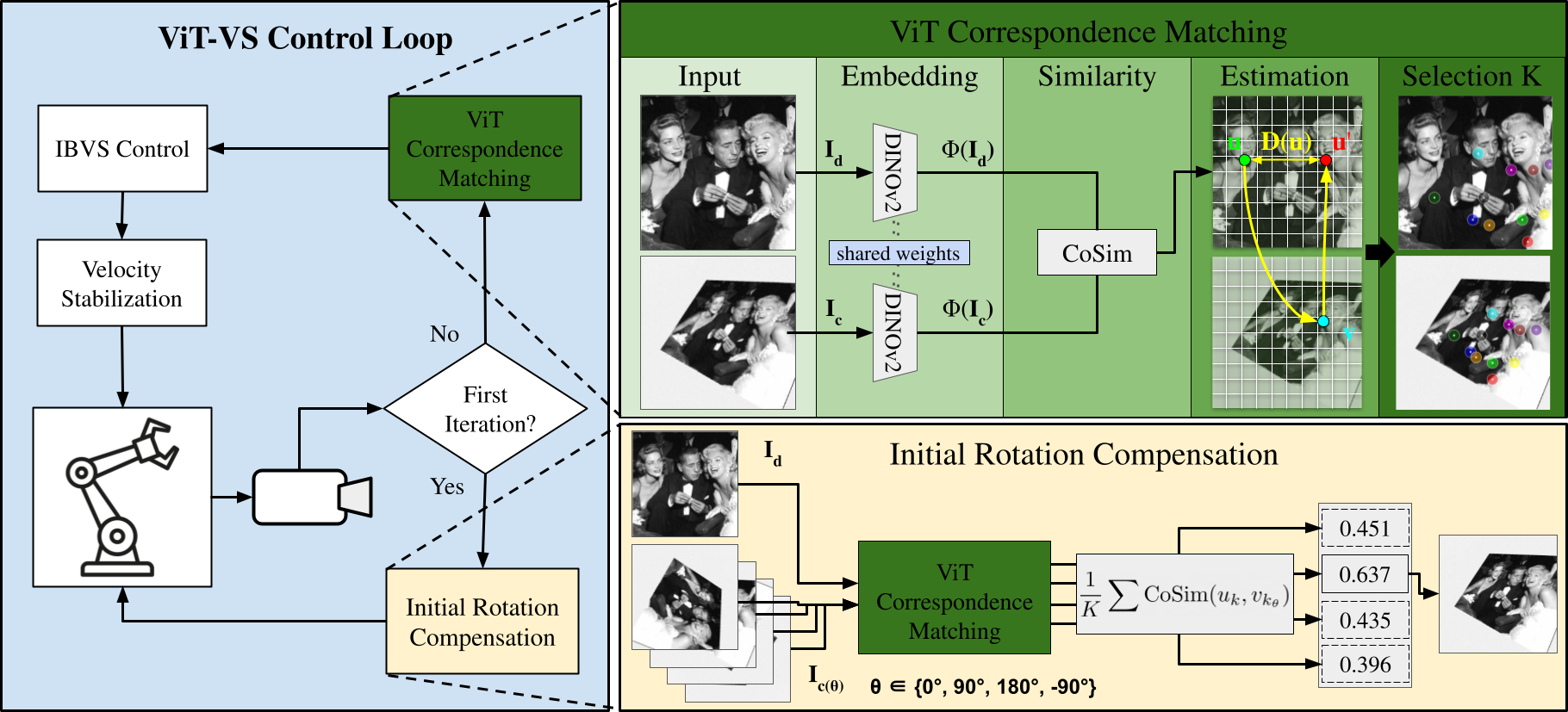}
    \caption{\textbf{ViT-VS Overview} Left: Visual servoing control loop integrating initial rotation compensation and IBVS control with ViT correspondences and velocity stabilization. Top right: Feature correspondence pipeline using DINOv2-based patch embeddings from desired and current image, where matches are estimated through cosine similarity and cyclical distance metrics, with random selection from top-K matches for spatial diversity. Bottom right: Initial rotation compensation mechanism evaluating four discrete rotations (0°, 90°, 180°, -90°) to determine optimal starting pose through mean feature similarity.}
    \label{fig:VitVS_overview}
    \vspace{-2ex}
\end{figure*}

\section{Deep Generalizable Visual Servoing}\label{sec:method}
Fig. \ref{fig:VitVS_overview} illustrates our deep zero-shot visual servoing pipeline, which combines ViT correspondence matching, initial rotation compensation, IBVS control and velocity stabilization.
Our approach leverages pretrained DINOv2~\cite{TMLR2023Oquab} models to extract patch embeddings. 
Correspondences are established using cosine similarity and cyclical distance metric.
To maintain spatial diversity across the image the correspondences are randomly selected from top-K matches.
To handle the inherent rotation invariance of ViTs, we implement an initial rotation compensation to evaluate the best initial pose.
The estimated patch correspondences are then used as input to the classical IBVS controller, with additional velocity stabilization through exponential moving average filtering to ensure smooth trajectories.
In the following sections, we describe each component of our pipeline in detail.

\subsection{ViT Correspondence Matching} \label{sec:correspondece_matching}
As visualized in the top right block of Fig.~\ref{fig:VitVS_overview} our method takes a desired image \( I_d \) and current image \( I_c \) as input.
Using the DINOv2 small model~\cite{TMLR2023Oquab}, we extract descriptor sets $\Phi(I_d), \Phi(I_c) \in \mathbb{R}^{H' \times W' \times D}$. 
Following \cite{CVPR2022Amir} and~\cite{ECCV2022Goodwin}, we adopt best buddy pairs matching concept from~\cite{CVPR2015Dekel}, which provides robust feature correspondences between images. 
Starting with a point $u \in \{1,\dots,H'\} \times \{1,\dots,W'\}$ in $\Phi(I_d)$, we find its nearest neighbor $v$ in $\Phi(I_c)$ using cosine similarity:
\begin{equation}
v = \underset{w}{\arg\max} \, \text{CoSim}(\Phi(I_d)_u, \Phi(I_c)_w)
\label{eq:nearest_neighbor1}
\end{equation}

To verify this match, we find the nearest neighbor $u'$ of $v$ back in $I_d$:
\begin{equation}
u' = \underset{z}{\arg\max} \, \text{CoSim}(\Phi(I_d)_z, \Phi(I_c)_v)
\label{eq:nearest_neighbor2}
\end{equation}

Using these matched pairs, we construct a cyclical distance map $D \in \mathbb{R}^{H' \times W'}$ as:
\begin{equation}
D_u = -\|u - u'\|_2
\label{eq:cyclical_distance}
\end{equation}

This cyclical distance as given in Equation \ref{eq:cyclical_distance} improves upon \cite{CVPR2015Dekel} by enabling the consistent selection of K semantic correspondences while incorporating spatial priors.
A cyclical distance of zero indicates a perfect match where the correspondence maps back to the original point.
We randomly select K correspondences that exceed our cyclical distance threshold, promoting better spatial distribution of features across the image.
This distribution strategy proves to be crucial for robust convergence by preventing feature concentration in visually distinctive regions.

\textbf{Feature Binning:}
To enhance feature robustness, we implement hierarchical feature binning as introduced by \cite{CVPR2022Amir}. The binning hierarchy parameter $\beta$ determines the contextual scope around each patch.
At $\beta=1$, each patch is combined with its eight immediate neighbors in a $3\times3$ grid.
Each increment of $\beta$ adds a new ring of context, with average pooling ensuring smooth feature transitions.
This hierarchical approach enriches the descriptors with broader contextual information at the cost of computational complexity.

\textbf{Operating Resolution:}
DINOv2 operates at input resolutions between $224\times224$ and $518\times518$ pixels.
Due to the computational demands of ViTs, feature extraction is limited to this resolution range.
Input images must be resized within this range, directly affecting both the granularity of extractable features and real-time performance capabilities.

\textbf{Foreground Segmentation}
In cases where the desired image includes background, it is necessary to utilize a foreground segmentation for the creation of segmentation masks.
For this purpose we utilize Segment Anything~\cite{kirillov2023segment}.

\subsection{Initial Rotation Compensation}
Vision Transformers, including DINOv2 \cite{TMLR2023Oquab}, are not inherently rotation invariant due to their architecture with fixed position encodings and patch-based processing \cite{ICLR2021Dosovitskiy}.
To address this limitation, we implement a rotation compensation step as shown in the bottom right of Fig. \ref{fig:VitVS_overview}. 

We evaluate the correspondence matching at four rotations (0°, 90°, 180°, -90°), calculating the mean cosine similarity score between the selected feature pairs $(u_k, v_k)$.
As shown in Equation \ref{eq:optimal_rotation}, we determine the optimal initial rotation angle $\theta^*$ that maximizes the mean similarity score in all matched pairs.
\begin{equation}
\theta^* = \underset{\theta \in \{0^{\circ}, 90^{\circ}, 180^{\circ}, -90^{\circ}\}}{\arg\max} \left(\frac{1}{K} \sum_{k=1}^K \text{CoSim}(u_k, v_k^\theta)\right)
\label{eq:optimal_rotation}
\end{equation}

This optimal rotation is applied to the robot, aligning the current image orientation with the desired image before initiating the visual servoing process.

\subsection{Image Based Visual Servoing with ViT}
Following rotation compensation, we implement classical IBVS as described by \cite{RAM2006Chaumette}.
The control law aims to minimize the error:
\begin{equation}
e(t) = s(m(t), a) - s^*
\label{eq:vs_error}
\end{equation}
where $s(m(t),a)$ represents the current image feature vector extracted from the image measurements $m(t)$ using camera intrinsic parameters $a$, and $s^*$ denotes the desired features. 
For $n$ feature points obtained through ViT matching, we compute the velocity control law as given in Equation \ref{eq:velocity_control}.

\begin{equation}
v_c = -\lambda \hat{L}_e^+ e
\label{eq:velocity_control}
\end{equation}

Here, $\hat{L}_e^+$ is the Moore-Penrose pseudoinverse of the interaction matrix, which we approximate using the depth values $Z_i$ of the corresponding feature points from the current depth-image.
The parameter $\lambda$ is included to ensure exponential error decrease.
For each feature point, the interaction matrix $L_i$ related to $s_i = (x_i, y_i)$ is constructed as:

\begin{equation}
L_i = \begin{bmatrix}
-\frac{1}{Z_i} & 0 & \frac{x_i}{Z_i} & x_iy_i & -(1+x_i^2) & y_i \\
0 & -\frac{1}{Z_i} & \frac{y_i}{Z_i} & 1+y_i^2 & -x_iy_i & -x_i
\end{bmatrix}
\label{eq:interaction_matrix}
\end{equation}

where $(x,y)$ are feature point coordinates transformed to real-world units using camera intrinsics. The complete interaction matrix combines these individual matrices as:

\begin{equation}
L_e = \begin{bmatrix} L_{1} \\ \vdots \\  L_{n-1} \\ L_{n} \end{bmatrix}
\label{eq:stacked_matrix}
\end{equation}

The control process follows three steps: first, calculating the error $e(t)$ as defined in Equation \ref{eq:vs_error}, then computing the interaction matrix $L_e$ as given in Equations \ref{eq:interaction_matrix} and \ref{eq:stacked_matrix}, and finally determining the velocity vector $v_c$ in Equation \ref{eq:velocity_control}.
The resulting velocity vector contains six components describing the cameras' linear and angular velocities in the camera coordinate frame.
\subsection{Velocity Stabilization} \label{sec:vel_stab}
Visual servoing systems face multiple uncertainty sources: feature detection variations, depth estimation inaccuracies, and numerical instabilities in matrix computations.
While our randomized feature selection improves convergence reliability, it can introduce velocity fluctuations affecting trajectory smoothness and mechanical stability.
To address this, we implement an Exponential Moving Average (EMA) filter for each velocity component:

\begin{equation}
v_t = \alpha v_{\text{new}} + (1-\alpha)v_{t-1}
\label{eq:ema_filter}
\end{equation}

where $v_t$ is the smoothed velocity, $v_{\text{new}}$ is the newly computed velocity, $v_{t-1}$ is the previous smoothed velocity, and $\alpha$ is the smoothing factor.
This filtering approach effectively dampens unwanted velocity fluctuations while maintaining system responsiveness, ensuring smooth trajectories throughout the servoing process.

\section{Experiments}\label{sec:experiments_results}
Our experimental evaluation consists of simulation studies, detailed system analysis, and extensive robotic experiments across different application scenarios.
We first describe our implementation details, followed by simulation experiments that benchmark our method against state-of-the-art approaches.
We then present system analysis results and conclude with real-world robotic experiments in three different scenarios.

\subsection{Experimental Setup}

\textbf{Hardware Setup} 
All experiments are conducted using a consistent hardware and software setup across simulation and real-world scenarios. 
The simulation experiments are conducted on a NVIDIA RTX 4060 Ti 16 GB and Intel Core i7-13700K CPU.
Robotic experiments are conducted on an NVIDIA RTX 4070 mobile GPU and AMD Ryzen 9 7940HS CPU.
The mobile manipulator platform consists of a Universal Robots UR5 mounted on a Mobile Industrial Robots MiR100. 
Industrial manipulation experiments are performed with a custom gripper and object grasping is performed with a Robotiq 2F-85.
An Intel RealSense D435i camera is used for all experiments.

\textbf{ViT Model Configuration} 
Our experimental setup employs DINOv2~\cite{TMLR2023Oquab}, pretrained on ImageNet1k~\cite{krizhevsky2012imagenet}, ViT-Small/14 architecture for feature extraction, using layer 11 for token-based features.
The model operates with a patch size and stride of $14$ pixels.
Feature binning is adopted from~\cite{CVPR2022Amir}.
Depending on the experiment, we use $\beta = 1$ or $\beta = 2$ binning hierarchies combined with a DINOv2 input resolution between $224\times224$ and $308\times308$ pixels as described in Section~\ref{sec:correspondece_matching}.
However, if not stated otherwise, we use $308\times308$ pixels and $\beta = 1$. 
For each iteration, we extract $24$ feature pairs for matching between current and desired view using the approach introduced in Section~\ref{sec:correspondece_matching}.
For similarity estimation and cyclic distance map computation we use the implementation of~\cite{ECCV2022Goodwin}, and for final correspondence selection we utilize a cutoff threshold of $1$. 

\textbf{Simulation Environment}
The simulation environment is replicated from Deep Metric Learning for Visual Servoing (DMLVS)~\cite{ICRA2023Felton}.
A virtual Intel RealSense D435i camera with a resolution of $640\times480$ pixels is used, the target is the \say{hollywood poster} in $60 \times 80 cm$, visible in Fig. \ref{fig:simulation_perturbation}.
We closely follow the approach of ~\cite{ICRA2023Felton} and generate $500$ distinct initial camera poses. 
Camera positions are sampled within a cuboid of $1.2m \times 1.2m \times 0.3m$ volume centered on the desired position.
The look-at points are distributed across four concentric circles on the poster plane, with radii of $8$, $16$, $24$, and $32 cm$ from the poster's center.
Camera orientations are samples using the look-at function, with an additional random rotation around the focal axis within [$-120\degree$, $120\degree$]. 
The desired camera position is set at $0.6m$ elevation from the poster center. 
This configuration yields average initial position errors of $46.42 \pm 16.99 cm$ and orientation errors of $74.12 \pm 27.71\degree$.

\textbf{Perturbation Settings}
To evaluate the robustness of our method and compare it to the deep learning results presented in~\cite{ICRA2023Felton} we conduct our experiments with unperturbed and perturbed images.
The perturbation parameters are taken from the codebase of~\cite{ICRA2023Felton}, and are implemented using Torchvision transforms: Colorjitter with a brightness of $0.6$ and contrast of $0.4$.
The random erasing with a probability of $0.5$ on a scale of $0.02$ to $0.33$ and a ratio of $0.3$ to $3.3$. The Gaussian blur is implemented with a mean of $0$ and a sigma of 0.05.

\textbf{Evaluation Metrics}
Convergence is reached when the velocities are close to zero, and initial position and rotation error are reduced by more than $90\%$, as done in~\cite{ICRA2023Felton}.
Furthermore we also report the Absolute Pose Error (APE) and length ratio;
APE quantifies the cumulated error in relation to the optimal PBVS trajectory and the length ratio quantifies the ratio between executed and ideal trajectory.

\subsection{Simulation Experiments}

\begin{figure}[t]
    \vspace{0.5ex}
    \centering
    \includegraphics[width=0.5\textwidth]{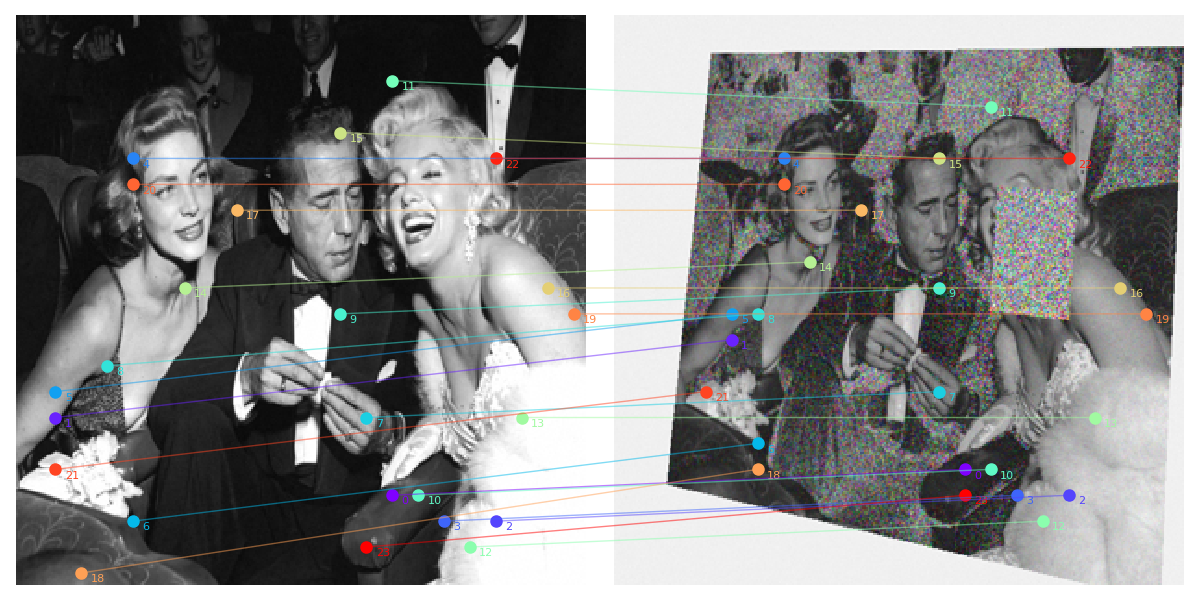}
    \caption{\textbf{ViT feature correspondences} between desired image (left) and current perturbed image (right) in simulation environment.}
    \label{fig:simulation_perturbation}
    \vspace{-2ex}
\end{figure}

This section compares ViT-VS with both classical and learning-based visual servoing approaches, in simulation.
To ensure fair comparison with classical ones, ViT-VS' matching strategy is replicated.
Feature matching with SIFT~\cite{lowe2004distinctive}, a floating-point descriptors, is done using the Euclidean distance, and for ORB~\cite{rublee2011orb} and AKAZE~\cite{alcantarilla2011fast}, both binary descriptors, the Hamming distance is used.
Matches are ranked by descriptor distance, and $24$ matches from the top-ranked candidates, as is the case for ViT-VS, are used for servoing.
Table \ref{tab:method_comparison} shows that ViT-VS converges with a success rate of $100\%$ in the unperturbed scenarios. 
Hence, the presented convergence rate is on par with the state-of-the-art learning-based DMLVS~\cite{ICRA2023Felton}, but without object- or scene-specific finetuning, and also significantly higher than that of classical descriptors.
Under image perturbations as exemplified in Fig. \ref{fig:simulation_perturbation}, ViT-VS achieves a $76.6\%$ success rate, improving over all classical feature-based methods and deep learning-based methods. 
The end error of ViT-VS, despite being inferior to classical methods, is competitive to PBVS approaches, in the case of image perturbations.
The translational end error is close to that of DMLVS, with $19.29\pm12.81cm$ for DMLVS and $21.54\pm12.11cm$ for ViT-VS, while the rotational end error is better, with $1.92\pm1.28\degree$ for DMLVS compared to $1.83\pm0.98\degree$ for ViT-VS.
The high end error in comparison to classical approaches is a consequence of the coarse feature maps of ViTs, which is $1/14$ of the input resolution in the case of our configuration with DINOv2-small.
This leads to correspondences being matched in a space with a resolution of $22\times22$, as compared to classical IBVS methods matching in a space with the resolution of the input image. 
Deep learned pose-based methods lead to a better APE and trajectory length since these metrics are designed for evaluating PBVS behavior.
ViT-VS achieves translational APE comparable to the best performing classical descriptor, ORB, $17.14\pm6.65cm$ compared to $16.60\pm5.66cm$, the best rotational accuracy $16.34\pm5.05\degree$ among non-finetuned methods, and the best length ratio at $1.21\pm0.39$.

Hence, ViT-VS combines the advantages of classical IBVS approaches, universal and general applicability since not requiring finetuning, and those of PBVS, high convergence rates, and better APE and length ratios than classical approaches. 
This makes our method ideal for real-world robotic manipulation tasks where generality is required, and consistent convergence outweighs sub-millimeter accuracy.

\begin{table*}[t!]
\vspace{1.5ex}
\caption{\textbf{Comparison to the state of the art} Results marked with an asterisk are taken from~\cite{ICRA2023Felton}.}
\label{tab:method_comparison}
\centering
\begin{tabular}{c|c|c|c|c|c|c|c|c}
\hline
\multirow{7}{*}[0em]{\begin{turn}{90}\textbf{Deep finetuning}\end{turn}} & \textbf{Method} & \textbf{Perturbed} & \textbf{Converged} & \textbf{End error} & \textbf{End error} & \textbf{APE} & \textbf{APE} & \textbf{Length} \\ 
& & \textbf{image} & \textbf{[\%]} & \textbf{[mm]} & \textbf{[°]} & \textbf{[cm]} & \textbf{[°]} & \textbf{ratio} \\ 
\cline{2-9}
& AEVS \cite{RAL2022Felton}* & $\times$ & $33.6$ & \underline{$0.01\pm0.00$} & \underline{$0.00\pm0.00$} & \underline{$2.74\pm6.36$} & $2.66\pm6.4$ & $2.75\pm4.85$ \\
\cline{2-9}
& PBVS-CNN, e.g., \cite{ICRA2018Bateux}* & $\times$ & $75.6$ & $33.52\pm6.45$ & $1.71\pm0.65$ & $3.13\pm1.04$ & $1.85\pm0.96$ & \underline{$1.11\pm0.08$} \\
& PBVS-CNN, e.g., \cite{ICRA2018Bateux}* & \checkmark & $36.8$ & $32.21\pm15.71$ & $2.37\pm1.57$ & $4.00\pm0.74$ & $\textbf{2.55}\pm\textbf{0.64}$ & $\textbf{1.12}\pm\textbf{0.10}$ \\
\cline{2-9}
& DMLVS, $K = 50$, \cite{ICRA2023Felton}* & $\times$ & \underline{$100.0$} & $0.04\pm0.03$ & \underline{$0.00\pm0.00$} & $4.00\pm0.72$ & \underline{$1.08\pm2.50$} & $1.18\pm0.23$ \\
& DMLVS, $K = 50$, \cite{ICRA2023Felton}* & \checkmark & $\textbf{76.0}$ & $\textbf{19.29}\pm\textbf{12.81}$ & $\textbf{1.92}\pm\textbf{1.28}$ & $\textbf{3.31}\pm\textbf{0.61}$ & $3.72\pm1.60$ & $1.14\pm0.11$ \\
\hline\hline
\multirow{13}{*}[1em]{\begin{turn}{90}\textbf{No finetuning}\end{turn}} & DVS \cite{ICRA2008Collewet}* & $\times$ & 9.8 & \underline{$0.00\pm0.00$} & \underline{$0.00\pm0.00$} & $17.32\pm12.48$ & $31.0\pm14.35$ & $2.60\pm4.90$ \\
\cline{2-9}
& SIFT IBVS & $\times$ & $89.6$ & $1.17\pm2.33$ & $0.09\pm0.11$ & $18.22\pm7.25$ & $27.76\pm11.31$ & $2.37\pm1.74$ \\
& SIFT IBVS & \checkmark & $24.0$ & $2.78\pm5.32$ & $0.23\pm0.47$ & $20.29\pm9.22$ & $26.66\pm11.15$ & $3.64\pm2.87$ \\
\cline{2-9}
& ORB IBVS & $\times$ & $98.6$ & $3.32\pm1.49$ & $0.25\pm0.12$ & \underline{$16.60\pm5.66$} & $25.66\pm10.64$ & $1.82\pm1.19$ \\
& ORB IBVS & \checkmark & $58.4$ & $3.86\pm3.36$ & $0.30\pm0.26$ & $\textbf{16.76}\pm\textbf{6.14}$ & $24.33\pm10.46$ & $1.91\pm1.18$ \\
\cline{2-9}
& AKAZE IBVS & $\times$ & $89.0$ & $1.03\pm1.21$ & $0.08\pm0.11$ & $16.99\pm5.90$ & $28.79\pm11.86$ & $1.91\pm1.26$ \\
& AKAZE IBVS & \checkmark & $58.0$ & $\textbf{1.44}\pm\textbf{1.06}$ & $\textbf{0.12}\pm\textbf{0.09}$ & $16.86\pm5.87$ & $25.70\pm10.71$ & $1.91\pm1.61$ \\
\cline{2-9}
& ViT-VS (no rot. comp.) & $\times$ & $83.8$ & $21.68\pm9.20$ & $1.66\pm0.69$ & $18.36\pm6.65$ & $25.37\pm9.61$ & $2.34\pm2.98$ \\
& ViT-VS (no rot. comp.) & \checkmark & $57.2$ & $24.10\pm11.80$ & $1.94\pm1.01$ & $17.94\pm6.38$ & $22.99\pm8.76$ & $2.95\pm3.40$ \\
\cline{2-9}
& \textbf{ViT-VS (ours)} & $\times$ & \underline{$100.0$} & $18.62\pm10.69$ & $1.50\pm0.78$ & $17.14\pm6.65$ & \underline{$16.34\pm5.05$} & \underline{$1.21\pm0.39$} \\
& \textbf{ViT-VS (ours)} & \checkmark & $\textbf{76.6}$ & $21.54\pm12.11$ & $1.83\pm0.98$ & $17.05\pm6.18$ & $\textbf{16.29}\pm\textbf{5.24}$ & $\textbf{1.90}\pm\textbf{1.44}$\\
\hline
\end{tabular}
    \vspace{-2ex}
\end{table*}

\textbf{Initial Rotation Compensation}
Table \ref{tab:method_comparison} shows that the rotation compensation improves ViT-VS performance significantly.
With this mechanism ViT-VS achieves $100.0\%$ and $76.6\%$, and without $83.8\%$ and $57.2\%$ convergence rates are achieved, for unperturbed and perturbed images, respectively.
The compensation allows our method to achieve state-of-the-art convergence rates while operating without object or scene-specific finetuning.

\begin{figure}[hbt]
    \centering
    \includegraphics[width=0.5\textwidth]{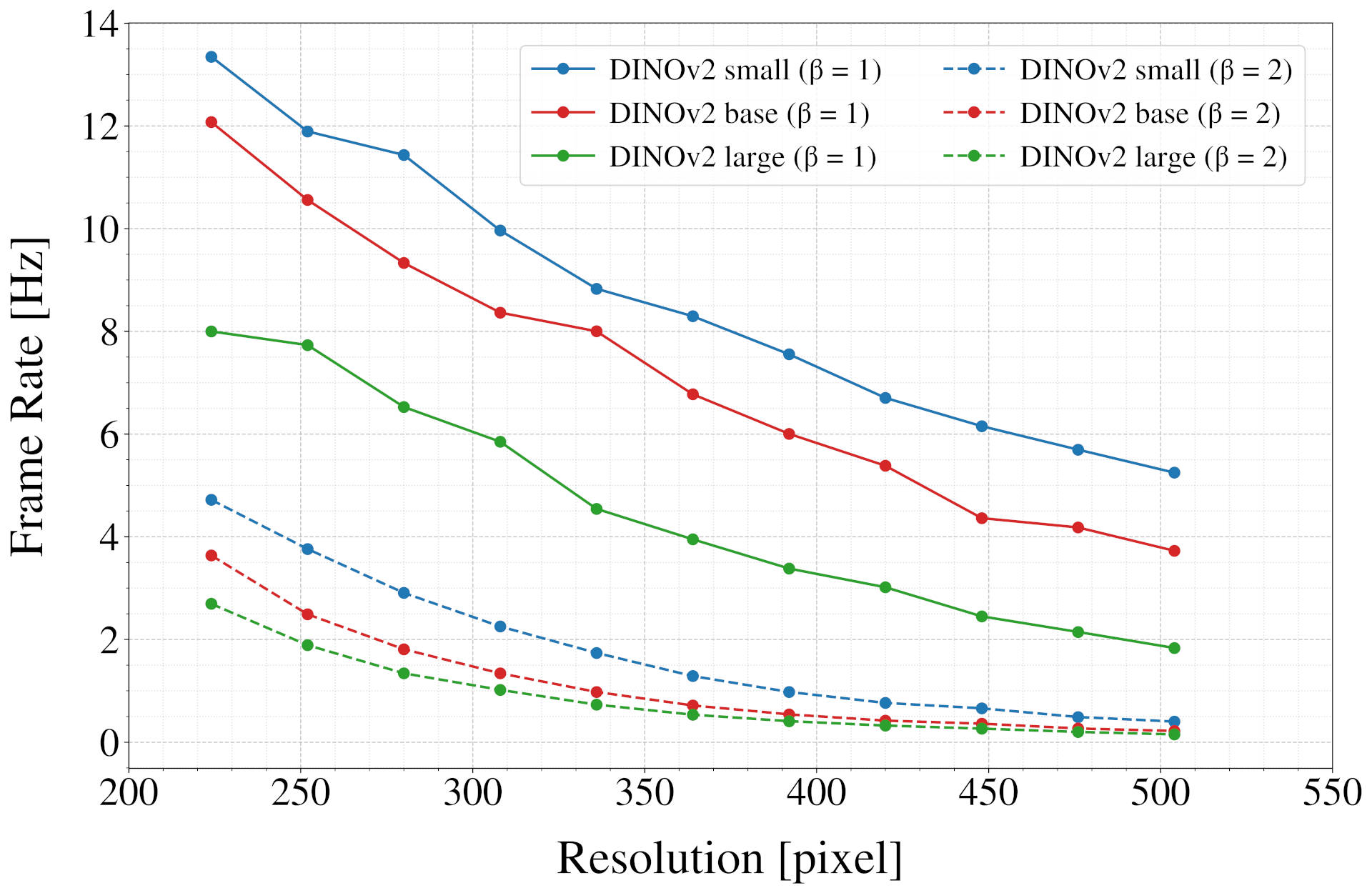}
    \caption{ViT-VS Frame Rate Analysis using NVIDIA RTX 4070 Mobile GPU across various model configurations.}
    \label{fig:runtime_graph}
\end{figure}

\textbf{Frame Rate Analysis}
In Fig. \ref{fig:runtime_graph} we presents the frame rate for ViT-VS, averaged over 100 runs using different configurations of DINOv2~\cite{TMLR2023Oquab}. 
Our experiments show that feature binning as introduced in Section \ref{sec:correspondece_matching} has a larger impact on computational efficiency than the choice of DINOv2 backbone size.
Based on these results, we focus on two configurations: DINOv2-Small with $224\times224$ pixel input and $\beta=2$ binning, or $308\times308$ pixel input with $\beta=1$ binning.
While the presented frame rates leave room for improvement, our robotic experiments show that the application of trajectory regularization effectively addresses motion jitter, resulting in smoother trajectories, as detailed in the following subsection.

\begin{figure}[hbt]
    \centering
    \includegraphics[width=0.99\columnwidth]{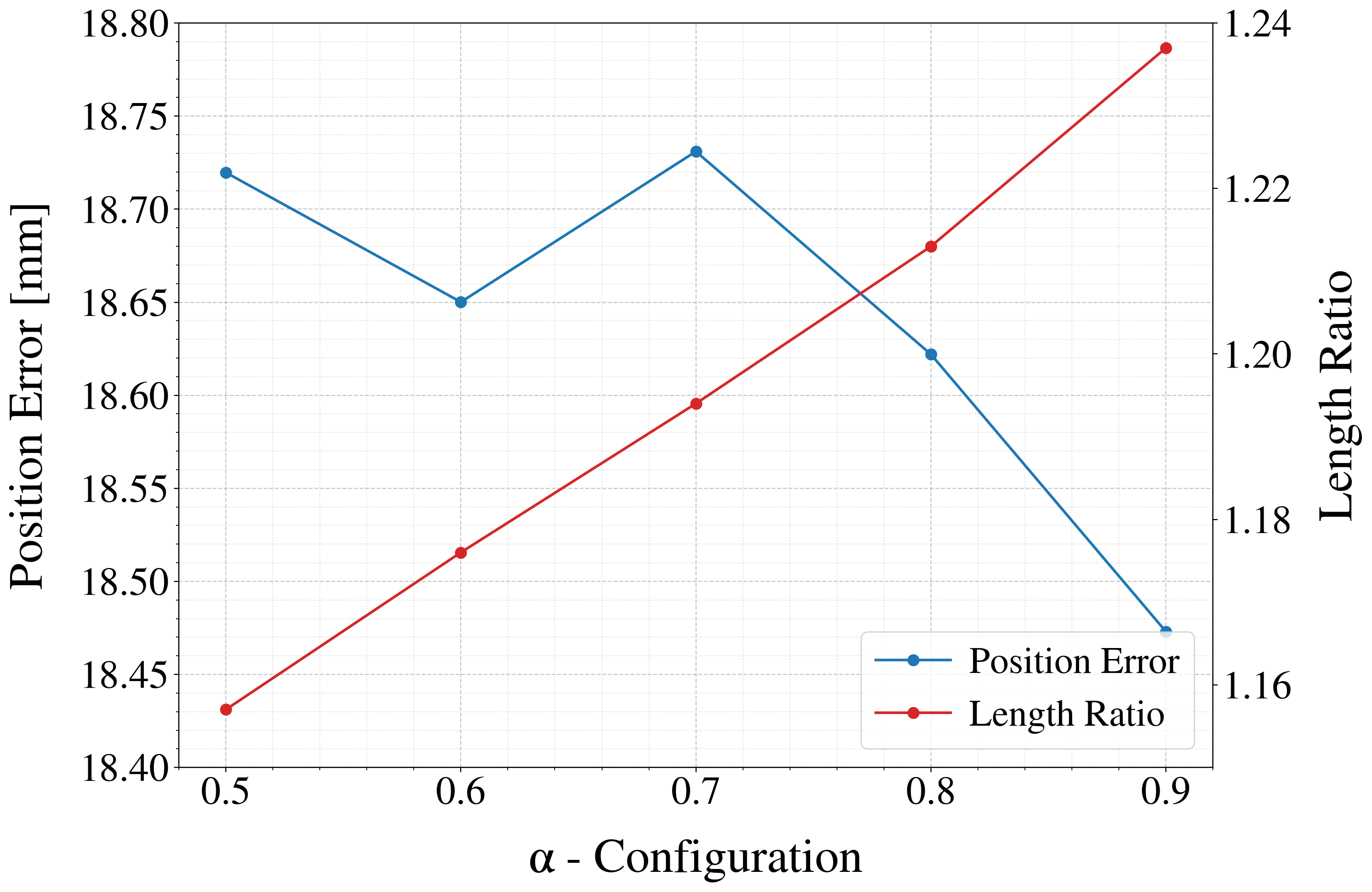}
    \caption{Alpha evaluation based on simulation runs without perturbation.}
    \label{fig:alpha_plot}
\end{figure}

\textbf{Trajectory Regularization}
The influence of the trajectory smoothing parameter $\alpha$ as defined in \ref{sec:vel_stab}, in a range from $0.5$ to $0.9$, is evaluated and illustrated in Fig. \ref{fig:alpha_plot}.
Lower $\alpha$ values lead to reduced length ratios, however, they also lead to an increase in end-positioning error.
We choose an $\alpha$ of $0.8$ as the standard configuration.
This value choice balances length ratio and end error.

\subsection{Robotic Experiments}

This section shows a detailed robotic experiment on the image that was used for comparing ViT-VS' convergence to the state of the art in simulation.  
Following that, experiments are provided that show that the presented method is suited for robustly compensating a mobile robot's positioning error for box manipulation.
Ultimately, we demonstrate that ViTs are semantically stable in a way to enable category-level object grasping.

\begin{figure}[!h]
    \vspace{1ex}
    \includegraphics[width=\columnwidth]{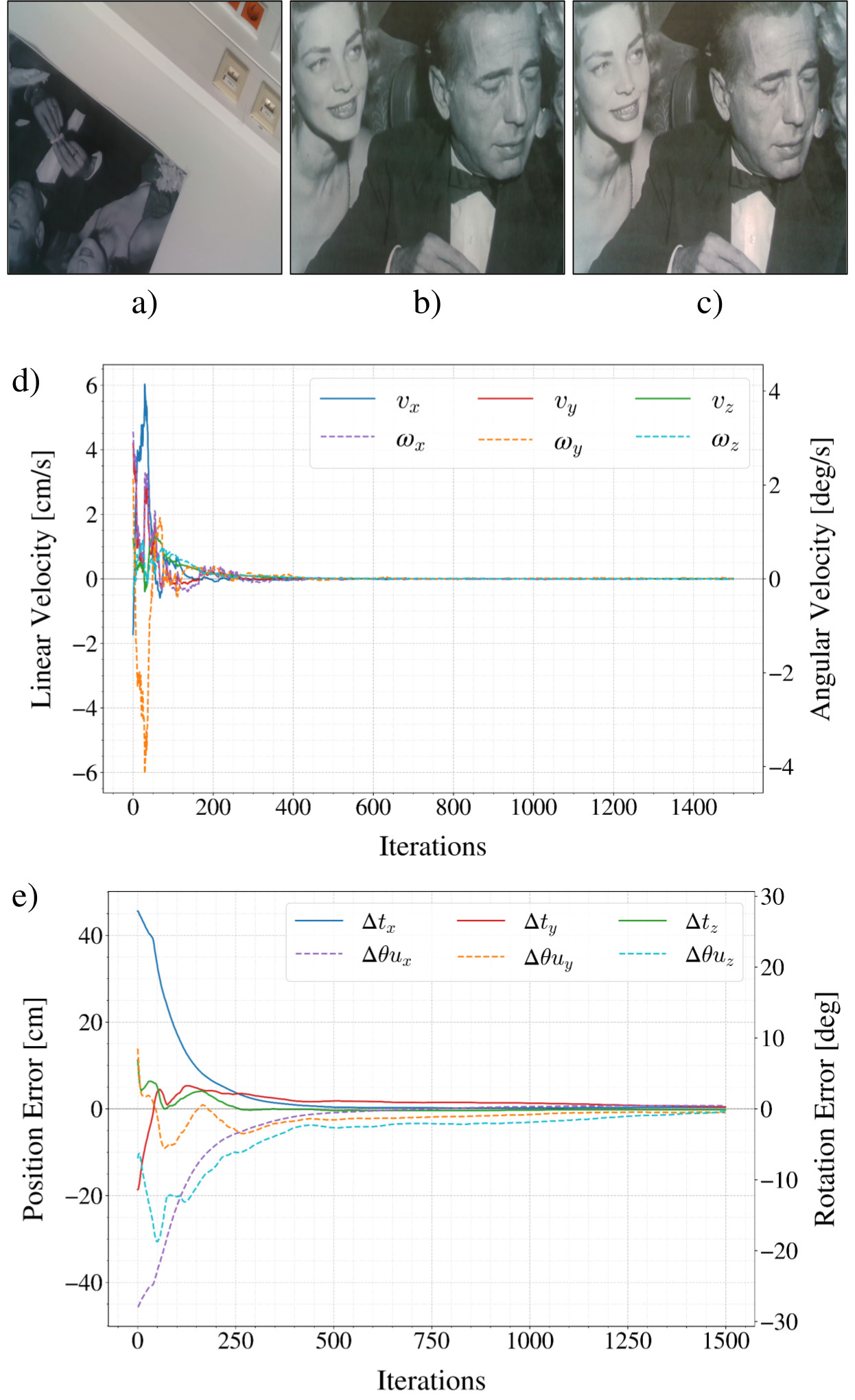}
    \caption{\textbf{Detailed robotic experiment} a) Initial image. b) Desired image. c) Final image. d) Camera velocities. e) Pose difference.}
    \label{fig:detailed-experiment}
\end{figure}

\textbf{Detailed Robotic Experiment}
This experiment demonstrates real-world evaluation on the \say{hollywood poster} for $1500$ iterations. 
The initial position error is 
$\Delta\mathbf{r_0} = (-45.60 cm, 18.63 cm, -11.21 cm,$
$10.17\degree, -15.48\degree, -153.08\degree)$.
Fig.~\ref{fig:detailed-experiment} visualizes the initial image (a), the desired image (b) and the final image (c).
Fig.~\ref{fig:detailed-experiment}(d) and (e) present the camera velocities and position and rotation errors.
The initial rotation compensation is not indicated since it is not part of the control loop. 
The best rotation was found to be a $180\degree$.
The final position error is $\Delta\mathbf{r_{final}} = (0.38 cm, 0.44 cm, -0.25 cm, 0.44\degree, -0.54\degree, -0.40\degree)$, showcasing full convergence from an only partially visible and heavily rotated initial position.

\begin{figure}[h!]
    \vspace{1ex}
    \includegraphics[width=\columnwidth]{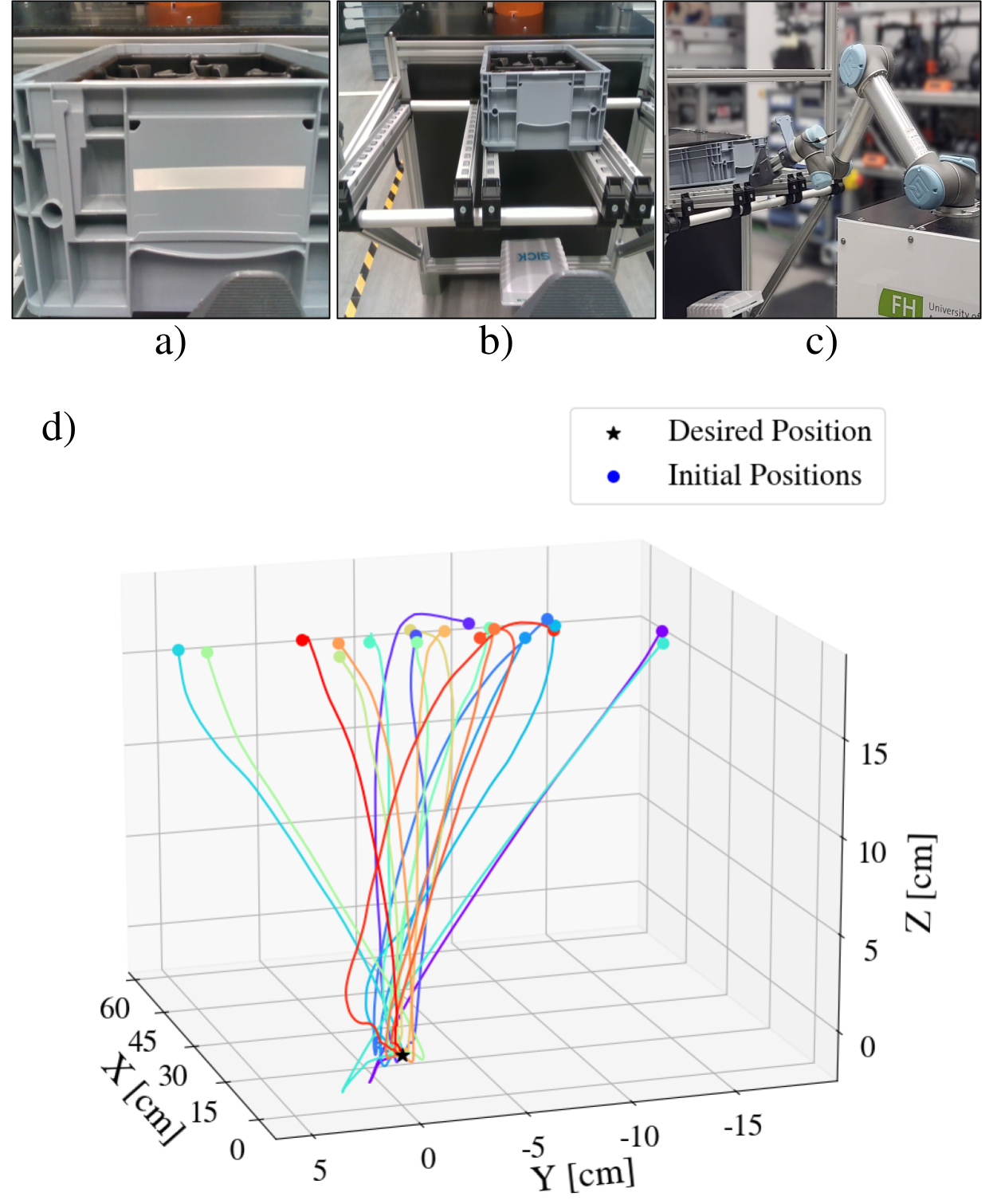}
    \caption{\textbf{Industrial use-case} a) Desired image. b) Example for initial image. c) External view before gripping. d) Tool center point trajectories plot aligned at goal position.}
    \label{fig:grab-box}
\end{figure}

\textbf{Industrial Use-case}
This experiment demonstrates the suitability of our approach for an industrial use case with a mobile manipulator. 
The mobile robot navigates to a work cell using ROS navigation stack~\cite{marder2010office} and two laser scanners.
This results in a positioning uncertainty of $\pm 10 cm$ of the MiR100 base.
Fig.~\ref{fig:grab-box} shows the desired image (a), an initial image (b), and the mobile robot after convergence (c). 
The trials are performed using boxes with different appearances, e.g., boxes with missing labels or structural differences.
We achieve a $100\%$ success rate on $20$ trials of box lifting.
The convergence behavior and positioning errors for all trials are visualized in Fig. \ref{fig:grab-box}(d). 
For this experiment $\beta=2$ and an image resolution of $224\times224$ pixels is used with DINOv2-Small to focus on geometry rather than texture.

\begin{figure}[h]
    \includegraphics[width=\columnwidth]{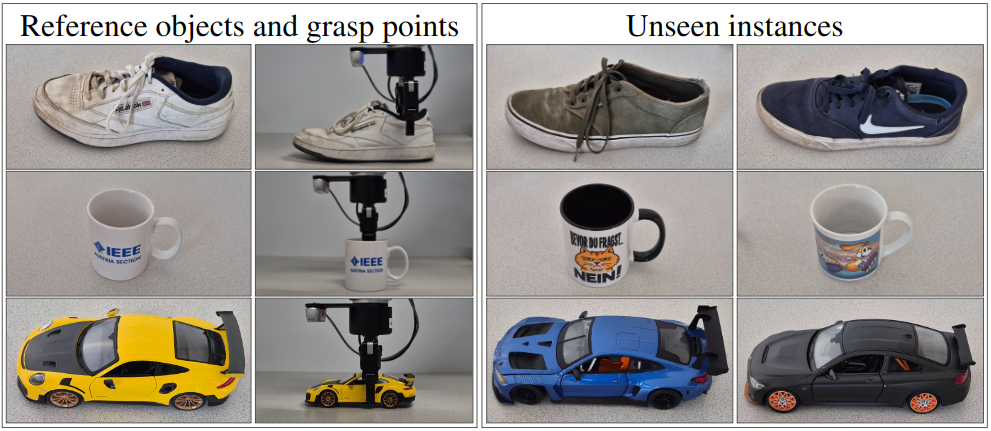}
    \caption{\textbf{Objects for sorting} Left side: Reference objects with corresponding grasp points. Right side: unseen object instances.}
    \label{fig:obstacles}
    \vspace{-2ex}
\end{figure}

\begin{table}[h!]
\caption{\textbf{Category-level object grasping experiments} Successes for grasping of singulated unseen object instances of the categories.}
\label{tab:category_level_robot}
\centering
\begin{tabular}{c|cc|cc|cc}
\hline
\textbf{Object} & \multicolumn{2}{c|}{\textbf{Car}} & \multicolumn{2}{c|}{\textbf{Shoe}}  & \multicolumn{2}{c}{\textbf{Mug}} \\ 
 & blue & black & blue & green & black & white \\ \hline 
Successes & \textbf{5/5} & 3/5 & \textbf{5/5} & \textbf{5/5} & 4/5 & \textbf{5/5} \\ %\hline
\hline
\end{tabular}
\end{table}

\begin{figure}[!h]
    \vspace{1.4ex}
    \includegraphics[width=\columnwidth]{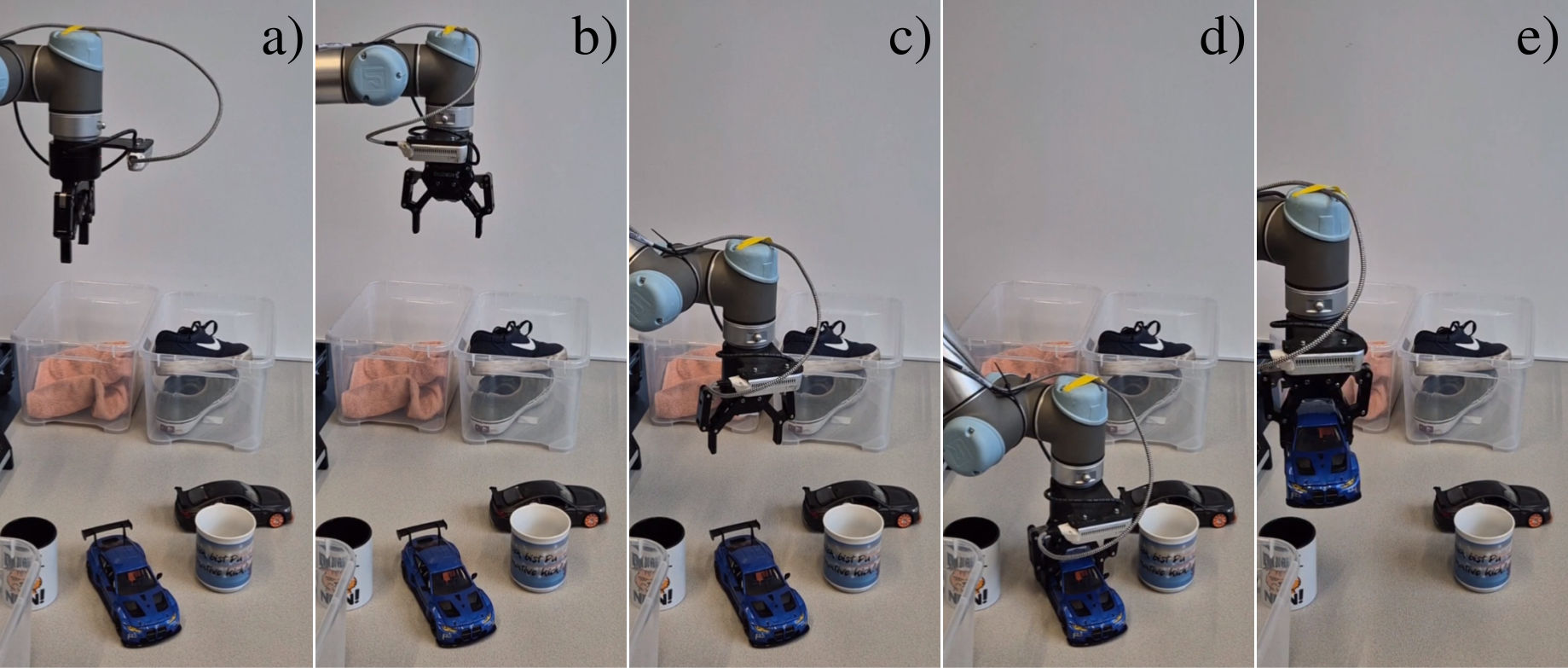}
    \caption{\textbf{Unseen object manipulation} Grasping of an unseen instance of the category toy car: a) initial position, b) compensated rotation, c) converged position, d) grasping, and e) object manipulation.}
    \label{fig:car_grasp}
    \vspace{-2ex}
\end{figure}

\textbf{Category-level Object Grasping}
Category-level object sorting experiments showcase the strong generalization capability of our method, allowing to perform pick and place tasks.
Table \ref{tab:category_level_robot} reports the success rates for grasping and sorting of randomly placed singulated objects.
The left part of Fig. \ref{fig:obstacles} shows the reference objects and the grasp points of the corresponding objects, the right part shows the unseen instance of the three categories mug, toy car, and shoe.
The left column shows the reference objects and the grasp points of the corresponding objects.
For each unseen object instance $5$ picking tries are performed; resulting in $10$ tries per object category.
The initial position is chosen to capture the full table plan.
Objects are separated from the background using Segment Anything~\cite{kirillov2023segment}.
ViT-VS, with a model configuration of $\beta=2$ and $224 \times 224$ input resolution, positions the end effector relative to the object for triggering the grasping motion, which is predefined for the seen object instance.
A successful grasp requires the robot to grasp and lift the object.
Our method demonstrates robust performance over all object categories, achieving success rates of $100\%$ for shoes, $90\%$ for mugs, and $80\%$ for toy cars.
The failed attempt for the mug occurred because the mug slipped out of the gripper after successful convergence and grasp.
The two failed attempts of the toy cars occurred due to table plane collisions; one due to bad convergence, one due to bad convergence caused by the initial rotation compensation retrieving the incorrect rotation.
Fig.~\ref{fig:car_grasp}(a)-(e) shows a representative grasping sequence of the blue unseen toy car.

\section{CONCLUSIONS}\label{sec:conclusion}

This work demonstrates the advantages of pretrained Vision Transformer features for visual servoing; Convergence rates are comparable to learning-based methods, yet features are generally applicable without finetuning, as is the case for classical image-based visual servoing.
Diverse robotics experiments demonstrate the usefulness and generality of Vision Transformer features for industrial tasks and household tasks, such as object manipulation and unseen object instance grasping.
Future work will investigate strategies for improving the positioning errors which are dictated by the resolution of the Vision Transformers' feature maps.

\addtolength{\textheight}{-12cm}

\bibliographystyle{IEEEtran} 
\bibliography{refs}

\end{document}